\documentclass[submission,copyright,creativecommons]{eptcs}
\providecommand{\event}{AREA 2022} 

\title{Towards Plug'n Play Task-Level Autonomy for Robotics Using POMDPs and
Generative Models}
\author{Or Wertheim,
\institute{Ben-Gurion University of the Negev}
\email{orwert@post.bgu.ac.il}
\and
Dan R.~Suissa \qquad\qquad 
\institute{Ben-Gurion University of the Negev}
\email{danrouve@bgu.ac.il}
\and
Ronen I. Brafman \qquad\qquad 
\institute{Ben-Gurion University of the Negev}
\email{brafman@cs.bgu.ac.il}
}
\def\titlerunning{Towards Plug'n Play Task-Level Autonomy for Robotics}
\def\authorrunning{Or Wertheim, Dan R.~Suissa, \& Ronen I. Brafman}

 \usepackage{graphicx}

 \usepackage{listings}
\usepackage{xcolor}
\colorlet{punct}{red!60!black}
\definecolor{background}{HTML}{EEEEEE}
\definecolor{delim}{RGB}{20,105,176}
\colorlet{numb}{magenta!60!black}
\lstdefinelanguage{json}{
    basicstyle=\normalfont\ttfamily,
    numbers=left,
    numberstyle=\scriptsize,
    stepnumber=1,
    numbersep=8pt,
    showstringspaces=false,
    breaklines=true,
    frame=lines,
    backgroundcolor=\color{background},
    literate=
     *{0}{{{\color{numb}0}}}{1}
      {1}{{{\color{numb}1}}}{1}
      {2}{{{\color{numb}2}}}{1}
      {3}{{{\color{numb}3}}}{1}
      {4}{{{\color{numb}4}}}{1}
      {5}{{{\color{numb}5}}}{1}
      {6}{{{\color{numb}6}}}{1}
      {7}{{{\color{numb}7}}}{1}
      {8}{{{\color{numb}8}}}{1}
      {9}{{{\color{numb}9}}}{1}
      {:}{{{\color{punct}{:}}}}{1}
      {,}{{{\color{punct}{,}}}}{1}
      {\{}{{{\color{delim}{\{}}}}{1}
      {\}}{{{\color{delim}{\}}}}}{1}
      {[}{{{\color{delim}{[}}}}{1}
      {]}{{{\color{delim}{]}}}}{1},
}
\usepackage{amsfonts}
\usepackage{amsmath}
\DeclareMathAlphabet{\mathcal}{OMS}{cmsy}{m}{n}
\begin{document}




\maketitle

\begin{abstract}
To enable robots to achieve high level objectives, engineers typically write scripts that  apply
existing specialized {\em skills}, such as navigation, object detection and manipulation to achieve these goals.  Writing good scripts is challenging since they must intelligently balance
the inherent stochasticity of a physical robot's actions and sensors, and the limited information it has. In principle, AI planning can be used to address this challenge and generate good behavior policies automatically. But this requires passing three hurdles. First, the AI must understand each skill's impact on the world. Second, we must bridge the gap between the
more abstract level at which we understand what a skill does
and the low-level state variables used within its code. Third, much integration effort is required to tie together all components.
We describe an approach for integrating robot {\em skills} into a working autonomous robot controller that
schedules its {\em skills} to achieve a specified
task and carries four key advantages. 
1) Our Generative Skill Documentation Language (GSDL)
makes code documentation simpler, compact, and more expressive
using ideas from probabilistic programming languages.
2) An expressive {\em abstraction mapping} (AM) bridges the gap between low-level robot code and the abstract AI planning model.
3) Any properly documented {\em skill} can be used by the controller without any additional programming effort, providing
a Plug'n Play experience. 4) A POMDP solver schedules skill execution
while properly balancing partial observability, stochastic behavior, and noisy sensing.
\end{abstract}


\section{Introduction}

To build autonomous robots capable of performing interesting tasks, one must integrate
multiple capabilities such as navigation, localization, different types of object manipulations, object detection, and more. Each of these areas attracts much research interest and our ability to program robots that can provide these capabilities, which we  refer to as
{\em skills}, has progressively improved. Moreover, for many skills, one can find publicly available software packages that implement them and publicly available algorithms one can implement independently. However, integrating diverse skills into a working system that can utilize them in unison to perform a given task is not easy.
First, this requires designing and implementing an execution system that can initiate the execution of each skill with the suitable parameters and adequately process the output of implemented sensing skills. 
Second, one must provide the logic that dictates which skills to use and when.
A solution must address both the software engineering challenge and
the conceptual issue of generating the execution logic, i.e., the behavior policy. 

The latter problem is often solved by manually writing a script.
Such pre-programmed scripts (can) have the advantage of being explainable and predictable. 
However, writing scripts for robotic agents is hard because physical agents' actions are usually probabilistic, and robots may have a partial noisy view of the world. Moreover, a script usually addresses a specific task only. To build autonomous systems that can perform diverse tasks in diverse environments, we must constantly supply new scripts or alter existing ones. For this reason, starting with the very
early days of robotics research, automated AI planning was suggested as a possible solution to the problem of generating a behavior policy~\cite{fikes1971strips}.

There is abundant work on the use of planning algorithms in robotics but these are mostly
one-of-a-kind implementations.
ROSPlan~\cite{cashmore2015rosplan} was one of the first systems attempting to address this issue by providing architecture and software that supports the integration of a planning engine into a ROS-based robot architecture~\cite{quigley2009ros}. Following ROSPlan, several other systems emerged that seek to
make the integration of planners into robot software easier, such as~\cite{niemueller2019goal,rovida2017skiros}. However, these systems have two main weaknesses.
First, they offer limited support for robots that operate with partial observability and use noisy sensors -- a basic property of many, if not most, mobile robotic systems. Second, they only partially address the integration issue discussed earlier, as they still require manual programming of standard interfaces between the skill's code and the engine. Moreover, they are often bound to specific systems, such as ROS. Finally, they rely on formal Action Description Languages (ADLs).

Indeed, most planning algorithms need, as input, an action
specifications in a formal language, such
as the Planning Domain Definition Language (PDDL)~\cite{fox2003pddl2}, the POMDP XML format (POMDPX) , etc.
Very few programmers are familiar with these languages and
it is difficult to specify stochastic effects and sensing
with them except in very small models\textsuperscript{\ref{AosVsRddlModel}}~\cite{silver2010monte}.
Instead, \cite{silver2010monte} is able to use a relatively simple, code-based generative model to model the game of Pacman, which has $10^{56}$ states. Their approach for modeling 
this large, nontrivial domain can be
divided into two. 1) Describe the planning domain via a 
sampling procedure, or simulator, that is able to sample
the next-state, the next-observation, and the next reward given the current state and action, correctly. A model that specifies how some object is
sampled, possibly dependent on some context parameters (e.g., a state and an action), is often called a generative model -- it shows how the object
is ``generated``. 2) Use code to describe this sampling procedure. Indeed, in the past decade or so, it was realized that programming languages could be adapted to serve as means of specifying complex generative models. This led
to the advent of probabilistic programming languages ~\cite{carpenter2017stan, ge2018turing} that, through the
use of code, can express complex generation processes and perform inference on them.

Code-based specification would not have worked with older planning algorithms that require ADL input. Yet, newer planners
based on sampling procedures, such as Partially Observable Monte-Carlo Planning (POMCP)~\cite{silver2010monte} and Determinized Sparse Partially Observable Tree (DESPOT)~\cite{somani2013despot} directly use code-based sampling procedures. Code-based specification is  typically better for such planners because
it can provide more efficient samplers than ones built from declarative   models\textsuperscript{\ref{AosVsRddl}}.

Among ADLs, the Relational Dynamic Influence Diagram Language (RDDL)~\cite{sanner2010relational} is noteworthy for its ability to compactly specify a generative probabilistic model using a dynamic Bayesian network model~\cite{ghahramani1997learning}.
Yet, it, too, does not have the expressiveness of programming languages, like advanced control structures (e.g., `while` loops) or built-in multi-purpose functions (e.g., C++ `cmath` library  
or the `string` class that provides string manipulation functions). 

A final crucial issue that requires attention is the abstraction gap. Action languages typically employ abstract concepts, such as {\em holding(cup)} or {\em at(kitchen)} to describe their model, whereas robotic code must interact with many lower-level variables. 

We seek to address existing systems' limitations and provide
programmers' with a plug'n play experience as follows:
The robot programmers program or import skills' code of their choice. They document their code using the more abstract Generative Skill Documentation Language (GSDL) and use an expressive {\em Abstraction Mapping} (AM) to bridge the gap between low-level robot code and the abstract AI planning model. Next, they need only supply a goal specification for each task and the system auto-generates all needed integration code and controls the robot online. 
The system described here\footnote{Our system's code is available at https://github.com/orhaimwerthaim/AOS-WebAPI/.} is part of the {\em Autonomous Robot Operating System} (AOS),
a general system we are developing for making programming of autonomous software from components easy.
This paper describes the decision engine of the system, which we will refer to as the AOS for brevity, despite its more limited scope.

The AOS can deal with partial observability and noisy sensing by using solution algorithms for partially observable Markov decision processes (POMDPs) and it uses ideas from probabilistic programming languages to make model specification easier and more flexible.
More specifically, our system makes the following contributions.
1. It introduces the Generative Skill Documentation Language (GSDL), a new code-based action description language
that supports stochastic actions and sensing and 
partial observability. 
2. It introduces a new {\em Abstraction Mapping} (AM) format that addresses the model-code abstraction gap.
3. It leverages the code in the GSDL to automatically generate efficient sampling code{\footnote{\label{AosVsRddl} An experiment~\cite{RDDLSimVsAOS} comparing sampling rates of  RDDLSim's~\cite{RDDLSim} generic code vs. AOS's domain-specific auto-generated code showed significant differences in favor of the AOS (452,000  vs. 12,500 samples per second).
}} for
sampling-based POMDP solvers and RL algorithms, but also supports ADL-based solvers.
4. 
It utilizes the knowledge in the AM to provide a  plug'n play experience in which code for integrating the planner and the diverse skills is auto-generated by the system, leaving the programmers with the sole task of describing their code and the task. 
5. Although currently demonstrated on ROS~\cite{quigley2009ros}, the architecture is general and can be 
converted for other robot frameworks.

Our empirical evaluation, involving different systems, demonstrates these capabilities, and its modular specification makes incremental development 
simple.

\section{Background and Related work}
We review POMDPs, AI planning architectures for robotics, 
and robot skills' documentation languages. 

\subsection{Partially Observable Markov Decision Process (POMDP)}
A discrete-time POMDP models the relationship between an agent and its environment. Formally, a POMDP is a tuple $\langle\mathcal{S},\mathcal{A},\mathcal{T},\mathcal{R},\Omega,\mathcal{O},\gamma, \mathcal{I}\rangle$: $\mathcal{S}$ is the state space, $\mathcal{A}$ is the action space, $\mathcal{T}$ is the state transition model,
$\mathcal{R}$ is the reward model,
 $\Omega$ is the observation space,
 $\mathcal{O}$ is the observation model, $\gamma \in [0,1]$ is the discount factor, and $\mathcal{I}\in \mathcal{B}$ is the initial belief state.  A belief state, which is a distribution
 over $\mathcal{S}$ is required since, in POMDPs the agent may not be fully aware of his current state.
 
 Following each action $a \in \mathcal{A}$, the environment
 transitions from its current state $s \in \mathcal{S}$ to
 state {\em s'} with probability $\mathcal{T}(s,a,s')$.
 Then, the agent receives an observation $o \in \Omega$, with probability $\mathcal{O}(s',a,o)$, and a reward $r=\mathcal{R}(a,s') \in	\mathbb{R}$.
 In the discounted case, we assume that earlier rewards are preferred and use a predefined discount factor $\gamma$ to reduce the utility of later rewards. The present value of a future reward {\em r} that
will be obtained at time {\em t} is hence $\gamma^tr$.
Using standard probabilistic inference, the updated belief
state $b'=Pr(s |a,o,b)$ can be computed from the model parameters.

A behavior {\em policy} for a POMDP, or simply a policy, is
a mapping $\pi:\mathcal{B} \mapsto \mathcal{A}$ from belief states to actions. The goal of POMDP solvers is to find a policy $\pi^*$ that maximize
the expected accumulated discounted reward, i.e.,
 $\pi^*=\underset{\pi}{max}[\mathbb{E}_\pi[\sum\nolimits_{t=1}^{\infty}\gamma^tr_t.]$.

POMDPs are a natural model for robots acting in the world because they capture the stochastic nature of robot's actions,
their noisy and partial sensing, and allow for diverse task specifications using the reward function.
  
\subsection{Planning-Based Deliberative Architectures}
Our work relates to deliberative robotic architectures, which follow the sense-plan-act paradigm, specifically those designed for general purpose rather than specific application.
In this respect, it includes the plan, act,
observe components discussed by~\cite{IngrandG17}.
The influential system that motivated much of our work is ROSPlan~\cite{cashmore2015rosplan}. ROSPlan is a planning and a plan execution architecture for robotics that generates
plans based on a PDDL2.1~\cite{fox2003pddl2} (or RDDL~\cite{sanner2010relational}) documentation of ROS-implemented skills. It supports a rich set of planning formalisms: classical, temporal, contingent planning, and probabilistic planning with some limitations~\cite{canal2019probabilistic}. However, even its
probabilistic variant maintains only a single world state that the user updates during execution, and it requires deterministic sensing.
When the inner state is discovered to be incorrect, the user can invoke replanning. As such, it cannot support full-fledged POMDP planning and cannot model the effect of  sensing actions on the belief state of the agent. 
Integration with ROSPlan requires  user effort~\cite{rosplanExp}, although recent work~\cite{rosplanBezrucav2021action} seeks to reduce it under certain conditions.  

The CLIPS Executive (CX)~\cite{niemueller2019goal} is a flexible robot execution and planning framework with some innovative ideas. It stores a predefined high-level plan in the form of a goal tree. CX calculates the next goal to pursue, and a PDDL solver generates a plan for this goal; based on the plan execution result, CX calculates the next goal and so on. The system preserves an extended model with the information required to activate robot skills. CX support for non-deterministic skills is limited to replanning. Nevertheless, it
proved its utility in a number of robotics competitions.%
\footnote{The platform was used by the winner of RoboCup German Open 2018 and PExC 2018.}
 
Unlike both systems, our system supports a full-fledged POMDP model and uses an expressive specification language.

SkiROS~\cite{rovida2017skiros} is a platform that can auto-generate action descriptions in PDDL based on a predefined ontology and invokes a solver to schedule the different robot skills. It includes a number of innovative ideas and a variety of tools. It, too, is based on classical planning with replanning, as opposed
to a POMDP model, 
and it requires users to work using strict patterns. Thus, code used must adapt to the architecture, whereas our system seeks to support integration of diverse code from diverse sources.

The system described in~\cite{lesire2020formalization}
and~\cite{AlboreDGLM21} proposes a formal language to specify robot skills with an expressive descriptive model used for reasoning, and an operational model. It maintains a life cycle for every robot skill and allows concurrent activation of the same skill. Code auto-generation assists users in integrating their code, yet users do need to add some code to handle changes in their skill life-cycle. This system uses a fixed policy described by an automaton or a behavior tree. Users can also use AI planning with PDDL solvers.  Our system does not support concurrent skill activation, but supports the richer POMDP model and code-based generative model specification. Moreover, it requires no additional information besides the documentation. 

\subsection{Skill Models}
Architectures that use a planner to control the execution of a set of skills require some form of skill documentation as
input to the planner. This documentation describes the effect of applying this skill/action on the system's state. 
ADLs such as STRIPS~\cite{fikes1971strips}, PDDL~\cite{younes2004ppddl1,fox2003pddl2} and RDDL~\cite{sanner2010relational} use formal syntax to describe
the action's effect. 
Most relevant to us, RDDL is  a language for describing dynamic Bayesian networks (DBNs)~\cite{ghahramani1997learning} that
is used for specifying transition and observation functions in MDPs and POMDPs.
It describes the post-action value of a state variable as a function of the pre-action variables' values. Moreover, RDDL allows the definition of intermediate effect variables for expressing more complex dependent effects.
RDDL specs, as well as their classical counterparts, can be understood as generative model specifications, as they implicitly describe how the post-action distribution is generated given pre-action values.
 Writing them, however, has some limitations:
a) RDDL syntax is less expressive than programming languages.
For example RDDL cannot describe a generative model that samples from a distribution until a condition is met since it does not support loops; 
b) probabilistic initial states are not supported;
c) hierarchical generative processes require intermediate variables definition, which may over-complicate the model.
GSDL, on the other end, has the expressive power of C++, which includes control structures and complex data structure
manipulation.
GSDL can easily describe real-world complex domains with probabilistic initial states, extrinsic changes, and action pre-condition. Each is in a designated area for a clear separation in the generative model. Moreover, it allows users to define hierarchical generative processes straightforwardly using code without intermediate variables.
Notably, the use of code, beyond making the specification process simpler, makes the sampling process required by the solver much more efficient\textsuperscript{\ref{AosVsRddl}} (the generative model itself is used for sampling). This translates into faster computation or (given similar time) better decisions.
%
The use of code (we support C++) also reduces
the amount of new syntax a programmer must master
to write a specification.





\section{System Overview and Concept} 
There is a long tradition of systems and architectures for autonomous robots 
based on tightly coupled components, such
as~\cite{Genome,skillman,rovida2017skiros} that provide various 
reasoning and planning services and provide support for programming skills in a principled manner.
Undoubtedly, such systems have shown some impressive results, 
yet while they offer various capabilities that can be exploited
when writing new code, such code must conform to the system's
requirements or methodology.

A more common approach with roots in the world of computer programming, is to try to re-use best-of-breed,
(or most-accessible) components, write additional functions/skills, and put them together. 
In robotics, we can use, for example, various ROS libraries, recent deep-learning-based object detection 
or object manipulation code, together with our own code for other needed skills. Our system, the AOS, takes this latter approach.%

\subsection{Concept}
The design process we support is the following: The user starts with a set of implemented skills, whether
imported or self-programmed. Each skill is a code module that can be activated and may respond with a returned value. These skills need to be documented. Code documentation is standard practice, but we require more formal documentation, consisting of two components, as described below. 
The {\em GSDL file}
describes how the execution of the code impacts the robot's and the world's state. 
The {\em Abstraction Mapping file} (AM) documents the connection between the abstract POMDP model depicted in the GSDL file and the skill code. The AM describes how to activate the code, how to map abstract parameter values to code-level parameters,
and how to compute the planning model-level observation
based on the robot skill execution output.
This provides a clean separation between the abstract system model captured by the GSDL file and low-level aspects captured by the AM file. An additional global {\em Environment file} is needed to specify the state variables, initial belief state, extrinsic changes, and special states (e.g., goal states). 

At this point, the user sends an HTTP request to the AOS Web API containing the path to their documented code. The AOS uses the GSDL and Environment files' code to auto-generate sampling code
that samples in accordance with the model specified in the GSDL fie.
The solver is then compiled and run. Similarly, a ROS {\em middleware node} that communicates with the solver is auto-generated based on the AM files. The robot and the {\em middleware node} are initialized, and an
online POMDP solver
now operates the robot, attempting to optimize its behavior. We use POMCP~\cite{silver2010monte}, but
any other online solver supporting the required API can be used. The user may query the AOS at any time for the execution status.
We also support the use of an off-line solver, desirable when the model is not too large and response times must be fast. For this purpose, we use the sampling code to convert the code-based generative model into a standard POMDP model and use the 
SARSOP solver~\cite{sarsop} to solve it.

The AOS auto-generates code for two purposes: 1) code required to run the POMDP solver that can sample states and observations using the GSDL files; 2) integration code, i.e., code that enables the solver to communicate with the skills, activate them, and receive `real-world` observations using the AM files.
This results in a true plug'n play experience: any executable skill on the robotic platform can be easily added to the system,
provided a GSDL and an associated AM file. Once added, the planning and execution engine can activate it with no additional effort.

\subsection{Skill Documentation}
The idea of using a formal description of an action as an input to a control algorithm underlies the area of AI planning~\cite{fikes1971strips,planning-book}, and goes
back to the robot Shakey~\cite{nilsson1984shakey}. Below we explain the language we use and its semantics. We start with the latter, explaining the generative model our documentation specifies, and then, through an example, we describe the
structure of our specification.

\subsubsection{Semantics and Structure}
Our specification describes a POMDP. Because our specification is code-based, this is not an explicit POMDP, but rather
an enhanced POMDP simulator. Enhanced because it contains information about the distributions from which state, observations, and rewards are sampled,
much like in probabilistic programming languages. We refer to it as a generative model because it explains how to generate the next state, observation, and reward from the current state and action.%
\footnote{The term {\em generative model} comes from the classification literature, while our models are dynamic, but it refers to models that specify the conditional probability of the observations given a class. 
That is, how the data collected is generated.}

Using code, we describe how the initial state is sampled and how the world changes. Changes occur in discrete steps (i.e., at this point, we ignore the duration of an action, although it can be used within the code), and can be exogenous or action induced. An action is selected at each time step. Before it is executed, an exogenous effect may take place. Then, the action is executed leading to a new state that depends on the state following
any exogenous event and the action. Depending on the resulting state and the action, an observation and a reward are received. 

The model specification is divided into multiple files. A global Environment File describes the POMDP elements unrelated to any specific robot skill: state variable definitions, initial belief state, 
the impact and likelihood of exogenous events, and state-dependent rewards. For each skill, a separate GSDL file documents the impact of that skill:
how it generates the next state, observation and reward, conditioned on the after exogenous effect state. This separation makes for a more manageable and 
 incremental software development process, and makes it  easy to export and continuously add documented skills. 

Each file has sections that correspond to the different elements it describes (e.g., initial state, observation probability, etc.). 
These sections contain sets of assignments that use C++ code lines.
In them, the modeler can refer to
three copies of state variables that can be conditioned on and assigned to: 1) the {\em previous state}, 2) the state {\em after extrinsic changes}, and 3) the {\em next state}. Moreover, there are variables for {\em met precondition}, {\em reward}, and {\em observation}. 

In addition, an AM file is associated with each skill, mapping between the skill's GSDL documentation to the skill code.

\subsubsection{Documentation Specification Through An Example} 
To illustrate actual documentation files, we describe part of the specification of  a toy problem. For more complete specification of the documentation format, see~\cite{aosExp}. 
In this problem, a robot with a single navigation skill must navigate as fast as possible to three known locations but we prefer that it will not visit the second location before visiting the first one.  The robot's initial location is unknown: it is the first location with probability 
0.5, and otherwise,most likely (80\%), it starts at the third location. Moreover, there is a 5\% chance that a person may occasionally move the robot, in which case it loses its orientation.

The navigation skill may fail, causing the robot to lose its orientation. Moreover, after experimenting with our navigation skill we know that: (1) Navigating the robot to its current location causes it to lose orientation. (2) It has a 10\% chance of losing its orientation while navigating to a different location. (3) The skill mistakenly reports success in 20\% of the cases in which the robot lost its orientation along the way. (4) When the robot loses orientation or starts navigating without knowing its location, the skill takes significantly longer to execute.

We describe abbreviated versions of the {\em Environment}, Navigation {\em GSDL}, and Navigation {\em AM} files for this example. In them, we distinguish between three values of each variable $x$.
Its value before  skill execution is denoted $state.x$. Its value after any extrinsic event is denoted
$state\_.x$. And its value after the skill execution is denoted $state\_\_.x$. State variables will also be referred to as {\em global variables} to distinguish them from {\em local variables}.
 
\paragraph{Environment File} 
Each robot has one {\em Environment file} that contains four sections.
1) The list of state variables (not shown) that comprise a POMDP state $s\in \mathcal{S}$. These may be primitive (e.g., int, string, bool, or float) or compound (custom types with sub-variables that are defined in
the Environment file) types. 2) A generative model of the initial belief state. Line 9 in Listing~\ref{lst:codeExampleEnv} describes the uncertainty regarding the robot's initial location.  
3) A generative model for extrinsic changes, possibly conditioned on the previous state. For example, a certain
constant probability of some malfunction when it is raining. Line 25 in Listing~\ref{lst:codeExampleEnv} describes the possible
effect of a person moving the robot. 4) An objective function as a set of state conditions and associated rewards. We can see in lines 11-21 in Listing~\ref{lst:codeExampleEnv} a high reward for visiting all locations that express our goal and a smaller negative reward to express our preference of not visiting the second location before visiting the first.

\paragraph{GSDL files}
A GSDL file is associated with a specific skill code and documents its expected behavior. 
It provides a quantitative description of the code's effects using concepts one would use to describe what one's code does in the world.
The GSDL file describes two elements of the global generative model: (1) Calculating the {\em met precondition} random variable. Lines 12-19 in Listing~\ref{lst:codeExamplePLP} express that we don't want the robot to navigate to its current location and lose its
orientation. (2) The dynamic model, i.e., how to sample the {\em next state}, action cost (or {\em reward}), and {\em observation} random variables. Lines 21-34 in Listing~\ref{lst:codeExamplePLP} describe it. 

Specifically, line 23 indicates that the robot loses orientation when navigating to its location or if the navigation fails (10\% chance), else it reaches its desired location. 
Line 27 defines the observation's generative model (called {\em moduleResponse}) to return a {\em Failed} observation 80\% of the time that the robot lost its orientation, expressing noisy sensing. Line 30 updates the {\em reward} model so that if the robot  \textbf{starts} navigating without knowing its location, it takes more time, expressed by a negative reward of minus five; otherwise, a function on the navigation distances expresses the time it takes to navigate. Finally, line 33 states a fixed large penalty when the navigation \textbf{ ends} in losing the orientation.  
Skills usually have parameters (e.g., destination of {\em move}), whose possible
values are currently defined in the Environment file. For example, in 
Lines 8-9 in Listing~\ref{lst:codeExamplePLP} the navigation destination 
is specified. 
The AOS Planning Engine instantiate any parameter with any legal parameter value when activating a skill.

\paragraph{Abstraction Mapping File (AM)} 
The AM file documents the abstract mapping between the robot code and the GSDL file (POMDP model) and serves as a bridge so the AOS can smoothly control the robot and reason about its execution outcomes. Each AM file is associated with the code for one skill and has two main roles that serve to map between code-level parameters and model-level parameters. 

The first role is to
describe how to activate the code. Lines 18-28 in Listing~\ref{lst:codeExampleGlue} describe how to activate a ROS service, specifying its path, service name, and parameters.  
The ROS service activation requires mapping high-level POMDP action parameters into lower-level code parameters, as defined in lines 43-54 in Listing~\ref{lst:codeExampleGlue} and used in line 25.

The second role is to define the observation associated with the skill execution outcome. Recall that in a POMDP an observation is obtained following each action execution. The AM computes the value of this observation from lower-level code parameters. Specifically, the AM specifies the
observation value, lines 6-17 in Listing Listing~\ref{lst:codeExampleGlue} describe the {\em Success} and {\em Failed} observations by referring to {\em local variables}. {\em Local variables} get their value in one of three ways. a) By a GSDL parameter. Lines 43-54 in Listing~\ref{lst:codeExampleGlue} define local variables for the `x,' `y,' `z' coordinates taken from the {\em desired location} GSDL parameter (line 8 in Listing~\ref{lst:codeExamplePLP}). b) As a function of public robot-framework data (e.g., ROS topics) or other local variables. Lines 36-42 in Listing~\ref{lst:codeExampleGlue} describe the {\em planSuccess} local variable, whose value is  {\em True} if during skill execution, the {\em /navigation/planner\_output} topic published a message containing the string 'success'. c) As a function of the skill code's response. Lines 30-35 in Listing~\ref{lst:codeExampleGlue} describe the {\em skillSuccess} local variable who stores the ROS service response.  

Thus, we see that the AM file can transform low-level public data into abstract observations correlated with the GSDL file and vice versa. 
AM files, like GSDL files, harness the expressive power of programming languages (the AM supports Python) to allow flexible integration with the AOS. Moreover, the user's sole work is to generate valid and coherent documentation, while the AOS supplies the tools to transform this documentation into a working autonomous robot. Furthermore, the AM allows more accurate reports of skill outcomes than initially coded and does so by reasoning with additional public data external to the skill code (lines 36-42 in Listing~\ref{lst:codeExampleGlue}).

\begin{lstlisting}[xleftmargin=7.0ex,xrightmargin=7.0ex,float,floatplacement=H,language=json,caption={{\em Environment} File Example.},label={lst:codeExampleEnv},basicstyle=\tiny,captionpos=b]
{
"GsdlMain": {
...
    "Type": "Environment"
}
...
"InitialBeliefStateAssignments": [
    {
        "AssignmentCode":  "state.robotLocation.discrete = AOS.Bernoulli(0.5) ? 1 (AOS.Bernoulli(0.2) ? 2 : 3);"
    }],
"SpecialStates": [
    {
        "StateConditionCode": "!state.v1.visited && state.v2.visited",
        "Reward": -50.0,
        "IsOneTimeReward": true
    },
    {
        "StateConditionCode": "state.v1.visited && state.v2.visited && state.v3.visited",
        "Reward": 7000.0,
        "IsGoalState": true
}],
"ExtrinsicChangesDynamicModel":
[
    {
        "AssignmentCode":  "if (AOS.Bernoulli(0.05)) state_.robotLocation.discrete = -1;"
    }
]}
\end{lstlisting}

\begin{lstlisting}[xleftmargin=7.0ex,xrightmargin=7.0ex,float,floatplacement=H,language=json,caption={Navigation Skill {\em GSDL} File Example.},label={lst:codeExamplePLP},basicstyle=\tiny,captionpos=b]
{
"GsdlMain": {...
    "Type": "GSDL"
    ...
},
"GlobalVariableModuleParameters": [
    {
        "Name": "oDesiredLocation",
        "Type": "tLocation"
    }
],
"Preconditions": {
    "GlobalVariablePreconditionAssignments": [
        {
            "AssignmentCode": "__meetPrecondition= oDesiredLocation.discrete != state.robotLocation.discrete;"
        }...
    ],
    "ViolatingPreconditionPenalty": -10
},
...
"NextStateAssignments": [
    {
        "AssignmentCode": " state__.robotLocation.discrete = !__meetPrecondition || AOS.Bernoulli(0.1) ? -1: oDesiredLocation.discrete;}"
    },
    ...
    {
        "AssignmentCode": "__moduleResponse = (state__.robotLocation.discrete == -1 && AOS.Bernoulli(0.8)) ? eFailed : eSuccess;"
    },
    {
        "AssignmentCode": "__reward = state_.robotLocation.discrete == -1 ? -5 : -(sqrt(pow(state.robotLocation.x-oDesiredLocation.x,2.0)+pow(state.robotLocation.y-oDesiredLocation.y,2.0)))*10;"
    },
    {
        "AssignmentCode": "if (state__.robotLocation.discrete == -1) __reward =  -10;"
    }
]
}
}
\end{lstlisting}

\begin{lstlisting}[xleftmargin=7.0ex,xrightmargin=7.0ex,float,floatplacement=H,language=json,caption={Navigation Skill {\em Abstraction Mapping} File Example.},label={lst:codeExampleGlue},basicstyle=\tiny,captionpos=b]
{
"GsdlMain": {
    ...
    "Type": "AM"
},
"ModuleResponse": {
    "ResponseRules": [
        {
            "Response": "eSuccess",
            "ConditionCodeWithLocalVariables": "skillSuccess and planSuccess"
        },
        {
            "Response": "eFailed",
            "ConditionCodeWithLocalVariables": "True"
        }
    ]
},
"ModuleActivation": {
    "RosService": {
        ...
        "ServicePath": "/navigate_to_point",
        "ServiceName": "navigate",
        "ServiceParameters": [
            { "ServiceFieldName": "goal",
               "AssignServiceFieldCode": "Point(x= nav_to_x, y= nav_to_y, z= nav_to_z)"}
        ]
    }
},
"LocalVariablesInitialization": [
    {
        "LocalVariableName": "skillSuccess",
        "FromROSServiceResponse": true,
        "AssignmentCode": "navigateSuccess=__input.success",
        ...
    },
    {
        "LocalVariableName": "planSuccess",
        "RosTopicPath": "/navigation/planner_output",
        "InitialValue": "False",
        ...
        "AssignmentCode": "if planSuccess == True:\n\treturn True\nelse:\n\treturn __input.data.find('success') > -1"
    }
    {
        "LocalVariableName": "nav_to_x",
        "FromGlobalVariable": "oDesiredLocation.x"
    },
    {
        "LocalVariableName": "nav_to_y",
        "FromGlobalVariable": "oDesiredLocation.y"
    },
    {
        "LocalVariableName": "nav_to_z",
        "FromGlobalVariable": "oDesiredLocation.z"
    }
]
}
\end{lstlisting}

\section{Experiments}
 We conducted several experiments~\cite{aosExp}
 to validate our system in different scenarios, described below.
 Their goal is to test ease of use, and the impact of relying on POMDP-based planners, and their highlights can be seen in our system overview video~\cite{aosOverviewVideo}). 


\subsection{TurtleBot3 Gazebo simulation}
Our first experiment used the TurtleBot3\cite{TurtleBot3} Gazebo simulation to see how we can quickly get
sophisticated behavior with little effort and existing code. The test environment included nine locations on a map. The goal was to
visit all locations while using a minimal length path. For navigation, we used ROS Move-Base~\cite{rosMoveBase}, restricted to start and end
positions that correspond to the nine locations. The programmer then defined a GSDL and AM files for this skill. The GSDL file uses
nine boolean variables that indicate whether a position was visited, a cost function that
is equal to distance travelled, and a reward for reaching all points. At this point, the AOS auto-generation code
generated the needed interfaces, and when the planner was activated, the robot performed the task, traveling the minimal distance.

\subsection{The Franka Emika Panda CoBot} 
Our second experiment involved a Panda CoBot~\cite{FrankaEmikaPanda} playing tic-tac-toe with a human (see video~\cite{aosPandaExpVideo}). An Intel RealSense D415 camera was
attached to the robot arm, and an erasable board with a tic-tac-toe grid was placed within its reach.
The experiment was based on two skills: marking a circle in a specific grid cell, and detecting change in the board state and extracting
the new board state. The first skill was implemented using our own PID controller based on \texttt{libfranka}, which we wrapped as a ROS service. The second skill was adapted from code found on the Web. 
After experimenting with the code to see its properties, GSDL and AM files were specified for each skill. 
The AOS allows the specification of an {\em Environment file} that describes exogenous events and are executed prior to every agent's action. This feature was used to model the human's action. We modeled the human as making random legal choices\footnote{\label{AosVsRddlModel}To model the human action, we used a C++ {\em while} loop that repeatedly sampled a tic-tac-toe cell until an empty one was sampled.
RDDL cannot compactly express behaviors of sampling until a condition is met. It would have to use an exhaustive {\em if-sample-else-sample} list which is only feasible for tiny distribution spaces.}.
Finally, we defined the goal reward, an initial state of empty board and the starting player, in the Environment file. 
Again, following the automated code generation, we  run the game (changing the
starting player, as desired). Because the human was modeled as a random player, you
can observe~\cite{aosPandaLostExpVideo} the robot sometimes relying on a human mistake of not completing a sequence of three.

\subsection{Armadillo Robot Gazebo Simulation}
The prior experiments involved mostly deterministic systems with full observability and few skills, and were aimed at showing the
plug'n play nature of the system. Our final experiment (see video~\cite{aosArmadilloVideo}) was conducted on a Gazebo simulation of our Armadillo robot
with more skills, partial observability, noisy sensing, and stochastic effects. These experiments demonstrate
the advantage of using a POMDP model, and the ease of incremental development (see \cite{aosArmadilloExp}).

The simulation environment included a room with two tables, and a corridor with a person. Each table had a can on it. One of the cans was very difficult to pick up (its true size was 10\% of the size perceived by the robot). The robot was located near the table with the difficult can. The goal was to give the can to a person in the corridor. Three skills were implemented by us: {\em pick-can, navigate}
which can navigate to a person, Table1, or Table 2, 
and {\em serve-can} which handed the can to the person. For the experiments, we used two versions
of the {\em pick} GSDL: a ``rough`` model that assumes that the probability of a successful pick action is independent
of the selected table, and a ``finer`` model in which the success probability is conditioned on the robot's position.

First, we experimented with each skill, saved statistics of their behavior, and used this information to write their GSDL files. 
In addition, we provided the AM files and the task specification. Again, this was sufficient to start the system going and attempt
to carry out the task. However, as the plan was executed, we saw that, occasionally, the pick skill ends with the arm outstretched.
Attempting to serve the person in this state causes a collision (i.e., injured the person). 
Moreover, {\em pick} returned success if motion planning and motion execution succeeded, but this did not imply that the can was
successfully picked up. Therefore, we wrote two new skills: {\em detect-hold-can} and {\em detect-arm-stretched}.
Implementing such skills that only map low-level public data (gripper pressure, arm-joint angles) to high-level insights is immediate. The user should only implement ROS services that do nothing and document them with GSDL and AM files. The AM files will describe the topics to listen to (e.g., gripper pressure, arm-joint angles) and their mappings to high-level observations. We also implemented an alternative approach where the sensing was integrated into the {\em pick} skill and
its return value now reflected the outcome of sensing whether the can
is held. This, too,
is very easy to do through the output specification in the AM file. 
Both involve small changes to the respective file.
{\em Detect-hold-can} is noisy and was modeled as such. 
{\em Detect-arm-stretched} is not noisy.  

First, with the rough model we saw, the robot (correctly) tries to pick the problematic can because it saves
the cost of navigating to the other table. With the finer model, it first moves to the other table where
pick action is more likely to succeed. Second, without the separated sensing actions, the robot serves the can, but then,
because it has no feedback, goes back to the tables and tries to repeat the process. With sensing, the robot
verifies success. If the result is yes, only then does it serve the can and stops. Moreover, since sensing
is noisy, the robot performs multiple sense actions to achieve a belief state with less uncertainty because
the results of the sensing actions are modeled as independent. However, when sensing is integrated into
the pick action, it cannot do independent sensing, and repeating the pick action is not desirable.



\begin{figure*}
    \centering
    \includegraphics[height=0.19\textwidth]{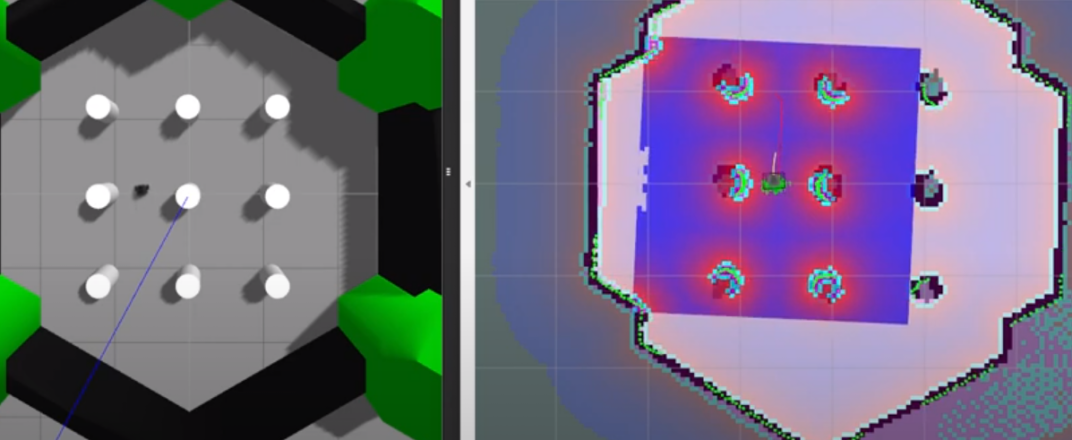} 
    \includegraphics[height=0.19\textwidth]{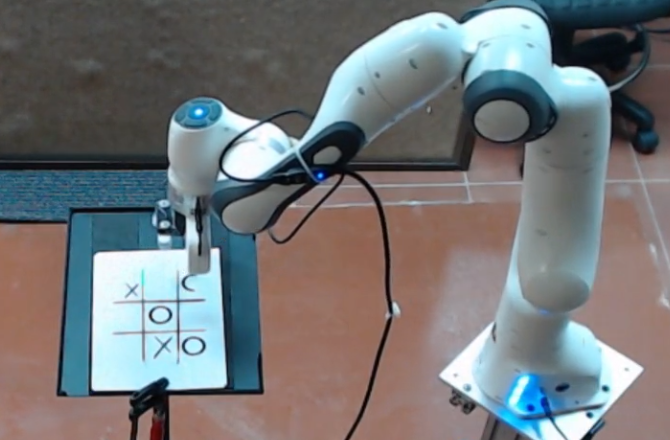}
    \includegraphics[height=0.19\textwidth]{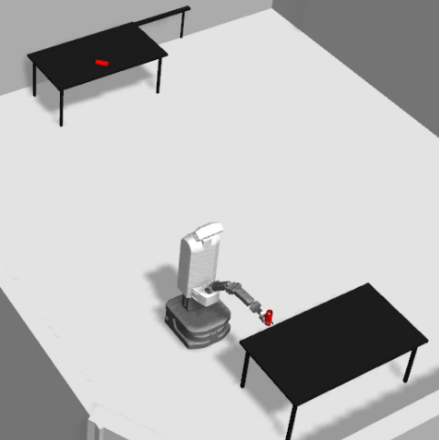}
    \caption{Experiments: (left) The TurtleBot3 Gazebo simulation and rviz. (center) The Franka Emika Panda. (right) The Armadillo Gazebo simulation. }

    \label{fig:foobar}
\end{figure*}

\section{Summary}
We presented the decision-engine of the AOS. 
Given a set of implemented skills, documented using a GSDL and AM files, the initial system state, and a reward specification,
the system  generates software that controls the robot by activating these skills as needed, taking care
of both execution logic and the software required to integrate all the components into a working system. Our empirical study demonstrated true plug'n play functionality and intelligent controller choices.

\subsection*{Acknowledgement}
This work was supported by the Ministry of Science and Technology's Grant \#3-15626,
by the Helmsley Charitable Trust through the Agricultural, Biological and Cognitive Robotics Center of Ben-Gurion University of the Negev, and the Lynn and William Frankel Center for Computer Science.

\bibliographystyle{eptcs}
\bibliography{bib}

\begin{thebibliography}{10}
\providecommand{\bibitemdeclare}[2]{}
\providecommand{\surnamestart}{}
\providecommand{\surnameend}{}
\providecommand{\urlprefix}{Available at }
\providecommand{\url}[1]{\texttt{#1}}
\providecommand{\href}[2]{\texttt{#2}}
\providecommand{\urlalt}[2]{\href{#1}{#2}}
\providecommand{\doi}[1]{doi:\urlalt{https://doi.org/#1}{#1}}
\providecommand{\eprint}[1]{arXiv:\urlalt{https://arxiv.org/abs/#1}{#1}}
\providecommand{\bibinfo}[2]{#2}

\bibitemdeclare{inproceedings}{AlboreDGLM21}
\bibitem{AlboreDGLM21}
\bibinfo{author}{Alexandre \surnamestart Albore\surnameend},
  \bibinfo{author}{David \surnamestart Doose\surnameend},
  \bibinfo{author}{Christophe \surnamestart Grand\surnameend},
  \bibinfo{author}{Charles \surnamestart Lesire\surnameend} \&
  \bibinfo{author}{Augustin \surnamestart Manecy\surnameend}
  (\bibinfo{year}{2021}): \emph{\bibinfo{title}{Skill-Based Architecture
  Development for Online Mission Reconfiguration and Failure Management}}.
\newblock In: {\slshape \bibinfo{booktitle}{3rd {IEEE/ACM} International
  Workshop on Robotics Software Engineering, RoSE@ICSE 2021, Madrid, Spain,
  June 2, 2021}}, \bibinfo{publisher}{{IEEE}}, pp. \bibinfo{pages}{47--54},
  \doi{10.1109/RoSE52553.2021.00015}.

\bibitemdeclare{inproceedings}{rosplanBezrucav2021action}
\bibitem{rosplanBezrucav2021action}
\bibinfo{author}{Stefan-Octavian \surnamestart Bezrucav\surnameend},
  \bibinfo{author}{Gerard \surnamestart Canal\surnameend},
  \bibinfo{author}{Michael \surnamestart Cashmore\surnameend} \&
  \bibinfo{author}{Burkhard \surnamestart Corves\surnameend}
  (\bibinfo{year}{2021}): \emph{\bibinfo{title}{An action interface manager for
  ROSPlan}}.
\newblock In: {\slshape \bibinfo{booktitle}{9th ICAPS Workshop on Planning and
  Robotics (PlanRob)}}, pp. \bibinfo{pages}{1751--1756},
  \doi{10.5281/zenodo.5348002}.

\bibitemdeclare{inproceedings}{canal2019probabilistic}
\bibitem{canal2019probabilistic}
\bibinfo{author}{Gerard \surnamestart Canal\surnameend},
  \bibinfo{author}{Michael \surnamestart Cashmore\surnameend},
  \bibinfo{author}{Senka \surnamestart Krivi{\'c}\surnameend},
  \bibinfo{author}{Guillem \surnamestart Aleny{\`a}\surnameend},
  \bibinfo{author}{Daniele \surnamestart Magazzeni\surnameend} \&
  \bibinfo{author}{Carme \surnamestart Torras\surnameend}
  (\bibinfo{year}{2019}): \emph{\bibinfo{title}{Probabilistic planning for
  robotics with ROSPlan}}.
\newblock In: {\slshape \bibinfo{booktitle}{Annual Conference Towards
  Autonomous Robotic Systems}}, \bibinfo{organization}{Springer}, pp.
  \bibinfo{pages}{236--250}, \doi{10.1007/978-3-030-23807-0\_20}.

\bibitemdeclare{article}{carpenter2017stan}
\bibitem{carpenter2017stan}
\bibinfo{author}{Bob \surnamestart Carpenter\surnameend},
  \bibinfo{author}{Andrew \surnamestart Gelman\surnameend},
  \bibinfo{author}{Matthew~D \surnamestart Hoffman\surnameend},
  \bibinfo{author}{Daniel \surnamestart Lee\surnameend}, \bibinfo{author}{Ben
  \surnamestart Goodrich\surnameend}, \bibinfo{author}{Michael \surnamestart
  Betancourt\surnameend}, \bibinfo{author}{Marcus \surnamestart
  Brubaker\surnameend}, \bibinfo{author}{Jiqiang \surnamestart Guo\surnameend},
  \bibinfo{author}{Peter \surnamestart Li\surnameend} \& \bibinfo{author}{Allen
  \surnamestart Riddell\surnameend} (\bibinfo{year}{2017}):
  \emph{\bibinfo{title}{Stan: A probabilistic programming language}}.
\newblock {\slshape \bibinfo{journal}{Journal of statistical software}}
  \bibinfo{volume}{76}(\bibinfo{number}{1}), \doi{10.18637/jss.v076.i01}.

\bibitemdeclare{inproceedings}{cashmore2015rosplan}
\bibitem{cashmore2015rosplan}
\bibinfo{author}{Michael \surnamestart Cashmore\surnameend},
  \bibinfo{author}{Maria \surnamestart Fox\surnameend}, \bibinfo{author}{Derek
  \surnamestart Long\surnameend}, \bibinfo{author}{Daniele \surnamestart
  Magazzeni\surnameend}, \bibinfo{author}{Bram \surnamestart
  Ridder\surnameend}, \bibinfo{author}{Arnau \surnamestart Carrera\surnameend},
  \bibinfo{author}{Narcis \surnamestart Palomeras\surnameend},
  \bibinfo{author}{Natalia \surnamestart Hurtos\surnameend} \&
  \bibinfo{author}{Marc \surnamestart Carreras\surnameend}
  (\bibinfo{year}{2015}): \emph{\bibinfo{title}{Rosplan: Planning in the robot
  operating system}}.
\newblock In: {\slshape \bibinfo{booktitle}{Twenty-Fifth International
  Conference on Automated Planning and Scheduling}}, pp.
  \bibinfo{pages}{1751--1756}, \doi{10.2478/CAIT-2012-0018}.

\bibitemdeclare{article}{skillman}
\bibitem{skillman}
\bibinfo{author}{Mohammed \surnamestart Diab\surnameend},
  \bibinfo{author}{Mihai \surnamestart Pomarlan\surnameend},
  \bibinfo{author}{Daniel \surnamestart Be{\ss}ler\surnameend},
  \bibinfo{author}{Aliakbar \surnamestart Akbari\surnameend},
  \bibinfo{author}{Jan \surnamestart Rosell\surnameend},
  \bibinfo{author}{John~A. \surnamestart Bateman\surnameend} \&
  \bibinfo{author}{Michael \surnamestart Beetz\surnameend}
  (\bibinfo{year}{2020}): \emph{\bibinfo{title}{SkillMaN - {A} skill-based
  robotic manipulation framework based on perception and reasoning}}.
\newblock {\slshape \bibinfo{journal}{Robotics Auton. Syst.}}
  \bibinfo{volume}{134}, p. \bibinfo{pages}{103653},
  \doi{10.1016/j.robot.2020.103653}.

\bibitemdeclare{manual}{FrankaEmikaPanda}
\bibitem{FrankaEmikaPanda}
\bibinfo{author}{Franka \surnamestart Emika\surnameend} (\bibinfo{year}{2021}):
  \emph{\bibinfo{title}{Franka Emika Panda cobot}}.
\newblock \urlprefix\url{{https://www.franka.de/robot-system/}}.

\bibitemdeclare{article}{fikes1971strips}
\bibitem{fikes1971strips}
\bibinfo{author}{Richard~E \surnamestart Fikes\surnameend} \&
  \bibinfo{author}{Nils~J \surnamestart Nilsson\surnameend}
  (\bibinfo{year}{1971}): \emph{\bibinfo{title}{STRIPS: A new approach to the
  application of theorem proving to problem solving}}.
\newblock {\slshape \bibinfo{journal}{Artificial intelligence}}
  \bibinfo{volume}{2}(\bibinfo{number}{3-4}), pp. \bibinfo{pages}{189--208},
  \doi{10.1016/0004-3702(71)90010-5}.

\bibitemdeclare{article}{fox2003pddl2}
\bibitem{fox2003pddl2}
\bibinfo{author}{Maria \surnamestart Fox\surnameend} \& \bibinfo{author}{Derek
  \surnamestart Long\surnameend} (\bibinfo{year}{2003}):
  \emph{\bibinfo{title}{PDDL2.1: An extension to PDDL for expressing temporal
  planning domains}}.
\newblock {\slshape \bibinfo{journal}{Journal of artificial intelligence
  research}} \bibinfo{volume}{20}, pp. \bibinfo{pages}{61--124},
  \doi{10.48550/arXiv.1106.4561}.

\bibitemdeclare{inproceedings}{ge2018turing}
\bibitem{ge2018turing}
\bibinfo{author}{Hong \surnamestart Ge\surnameend}, \bibinfo{author}{Kai
  \surnamestart Xu\surnameend} \& \bibinfo{author}{Zoubin \surnamestart
  Ghahramani\surnameend} (\bibinfo{year}{2018}): \emph{\bibinfo{title}{Turing:
  a language for flexible probabilistic inference}}.
\newblock In: {\slshape \bibinfo{booktitle}{International conference on
  artificial intelligence and statistics}}, \bibinfo{organization}{PMLR}, pp.
  \bibinfo{pages}{1682--1690}, \doi{10.17863/CAM.42246}.

\bibitemdeclare{inproceedings}{ghahramani1997learning}
\bibitem{ghahramani1997learning}
\bibinfo{author}{Zoubin \surnamestart Ghahramani\surnameend}
  (\bibinfo{year}{1997}): \emph{\bibinfo{title}{Learning dynamic Bayesian
  networks}}.
\newblock In: {\slshape \bibinfo{booktitle}{International School on Neural
  Networks, Initiated by IIASS and EMFCSC}}, \bibinfo{organization}{Springer},
  pp. \bibinfo{pages}{168--197}, \doi{10.1007/BFb0053999}.

\bibitemdeclare{book}{planning-book}
\bibitem{planning-book}
\bibinfo{author}{Malik \surnamestart Ghallab\surnameend},
  \bibinfo{author}{Dana~S. \surnamestart Nau\surnameend} \&
  \bibinfo{author}{Paolo \surnamestart Traverso\surnameend}
  (\bibinfo{year}{2016}): \emph{\bibinfo{title}{Automated Planning and
  Acting}}.
\newblock \bibinfo{publisher}{Cambridge University Press},
  \doi{10.1017/CBO9781139583923}.

\bibitemdeclare{article}{IngrandG17}
\bibitem{IngrandG17}
\bibinfo{author}{F{\'{e}}lix \surnamestart Ingrand\surnameend} \&
  \bibinfo{author}{Malik \surnamestart Ghallab\surnameend}
  (\bibinfo{year}{2017}): \emph{\bibinfo{title}{Deliberation for autonomous
  robots: {A} survey}}.
\newblock {\slshape \bibinfo{journal}{Artif. Intell.}} \bibinfo{volume}{247},
  pp. \bibinfo{pages}{10--44}, \doi{10.1016/j.artint.2014.11.003}.

\bibitemdeclare{inproceedings}{lesire2020formalization}
\bibitem{lesire2020formalization}
\bibinfo{author}{Charles \surnamestart Lesire\surnameend},
  \bibinfo{author}{David \surnamestart Doose\surnameend} \&
  \bibinfo{author}{Christophe \surnamestart Grand\surnameend}
  (\bibinfo{year}{2020}): \emph{\bibinfo{title}{Formalization of robot skills
  with descriptive and operational models}}.
\newblock In: {\slshape \bibinfo{booktitle}{2020 IEEE/RSJ International
  Conference on Intelligent Robots and Systems (IROS)}},
  \bibinfo{organization}{IEEE}, pp. \bibinfo{pages}{7227--7232},
  \doi{10.1109/IROS45743.2020.9340698}.

\bibitemdeclare{inproceedings}{Genome}
\bibitem{Genome}
\bibinfo{author}{Anthony \surnamestart Mallet\surnameend},
  \bibinfo{author}{C{\'{e}}dric \surnamestart Pasteur\surnameend},
  \bibinfo{author}{Matthieu \surnamestart Herrb\surnameend},
  \bibinfo{author}{S{\'{e}}verin \surnamestart Lemaignan\surnameend} \&
  \bibinfo{author}{Fran{\c{c}}ois~Felix \surnamestart Ingrand\surnameend}
  (\bibinfo{year}{2010}): \emph{\bibinfo{title}{GenoM3: Building
  middleware-independent robotic components}}.
\newblock In: {\slshape \bibinfo{booktitle}{{IEEE} International Conference on
  Robotics and Automation, {ICRA} 2010, Anchorage, Alaska, USA, 3-7 May 2010}},
  \bibinfo{publisher}{{IEEE}}, pp. \bibinfo{pages}{4627--4632},
  \doi{10.1109/ROBOT.2010.5509539}.

\bibitemdeclare{manual}{rosMoveBase}
\bibitem{rosMoveBase}
\bibinfo{author}{Eitan \surnamestart Marder-Eppstein\surnameend}
  (\bibinfo{year}{2021}): \emph{\bibinfo{title}{ROS Move-Base}}.
\newblock \urlprefix\url{{http://wiki.ros.org/move_base}}.

\bibitemdeclare{inproceedings}{niemueller2019goal}
\bibitem{niemueller2019goal}
\bibinfo{author}{Tim \surnamestart Niemueller\surnameend},
  \bibinfo{author}{Till \surnamestart Hofmann\surnameend} \&
  \bibinfo{author}{Gerhard \surnamestart Lakemeyer\surnameend}
  (\bibinfo{year}{2019}): \emph{\bibinfo{title}{Goal reasoning in the CLIPS
  Executive for integrated planning and execution}}.
\newblock In: {\slshape \bibinfo{booktitle}{Proceedings of the International
  Conference on Automated Planning and Scheduling}}, \bibinfo{volume}{29}, pp.
  \bibinfo{pages}{754--763}.

\bibitemdeclare{article}{nilsson1984shakey}
\bibitem{nilsson1984shakey}
\bibinfo{author}{Nils~J \surnamestart Nilsson\surnameend}
  (\bibinfo{year}{1984}): \emph{\bibinfo{title}{Shakey the robot}}.
\newblock {\slshape \bibinfo{journal}{Institute for Software Technology}}.

\bibitemdeclare{inproceedings}{quigley2009ros}
\bibitem{quigley2009ros}
\bibinfo{author}{Morgan \surnamestart Quigley\surnameend}, \bibinfo{author}{Ken
  \surnamestart Conley\surnameend}, \bibinfo{author}{Brian \surnamestart
  Gerkey\surnameend}, \bibinfo{author}{Josh \surnamestart Faust\surnameend},
  \bibinfo{author}{Tully \surnamestart Foote\surnameend},
  \bibinfo{author}{Jeremy \surnamestart Leibs\surnameend}, \bibinfo{author}{Rob
  \surnamestart Wheeler\surnameend} \& \bibinfo{author}{Andrew~Y \surnamestart
  Ng\surnameend} (\bibinfo{year}{2009}): \emph{\bibinfo{title}{ROS: an
  open-source Robot Operating System}}.
\newblock In: {\slshape \bibinfo{booktitle}{ICRA workshop on open source
  software}}, \bibinfo{volume}{3.2}, \bibinfo{organization}{Kobe, Japan},
  p.~\bibinfo{pages}{5}.

\bibitemdeclare{manual}{TurtleBot3}
\bibitem{TurtleBot3}
\bibinfo{author}{\surnamestart ROBOTIS\surnameend}:
  \emph{\bibinfo{title}{TurtleBot3 e-Manual}}.
\newblock
  \urlprefix\url{https://emanual.robotis.com/docs/en/platform/turtlebot3/overview/}.

\bibitemdeclare{incollection}{rovida2017skiros}
\bibitem{rovida2017skiros}
\bibinfo{author}{Francesco \surnamestart Rovida\surnameend},
  \bibinfo{author}{Matthew \surnamestart Crosby\surnameend},
  \bibinfo{author}{Dirk \surnamestart Holz\surnameend},
  \bibinfo{author}{Athanasios~S \surnamestart Polydoros\surnameend},
  \bibinfo{author}{Bjarne \surnamestart Gro{\ss}mann\surnameend},
  \bibinfo{author}{Ronald \surnamestart Petrick\surnameend} \&
  \bibinfo{author}{Volker \surnamestart Kr{\"u}ger\surnameend}
  (\bibinfo{year}{2017}): \emph{\bibinfo{title}{SkiROS—a skill-based robot
  control platform on top of ROS}}.
\newblock In: {\slshape \bibinfo{booktitle}{Robot operating system (ROS)}},
  \bibinfo{publisher}{Springer}, pp. \bibinfo{pages}{121--160},
  \doi{10.1007/978-3-319-54927-9\_4}.

\bibitemdeclare{unpublished}{RDDLSim}
\bibitem{RDDLSim}
\bibinfo{author}{Scott \surnamestart Sanner\surnameend} (\bibinfo{year}{2010}):
  \emph{\bibinfo{title}{Implements a parser, simulator, and client/server
  evaluation architecture for the relational dynamic influence diagram language
  (RDDL)}}.
\newblock \bibinfo{note}{Https://github.com/ssanner/rddlsim}.

\bibitemdeclare{article}{sanner2010relational}
\bibitem{sanner2010relational}
\bibinfo{author}{Scott \surnamestart Sanner\surnameend} (\bibinfo{year}{2010}):
  \emph{\bibinfo{title}{Relational dynamic influence diagram language (RDDL):
  Language description}}.
\newblock {\slshape \bibinfo{journal}{Unpublished ms. Australian National
  University}} \bibinfo{volume}{32}, p.~\bibinfo{pages}{27}.

\bibitemdeclare{inproceedings}{silver2010monte}
\bibitem{silver2010monte}
\bibinfo{author}{David \surnamestart Silver\surnameend} \&
  \bibinfo{author}{Joel \surnamestart Veness\surnameend}
  (\bibinfo{year}{2010}): \emph{\bibinfo{title}{Monte-Carlo planning in large
  POMDPs}}.
\newblock In: {\slshape \bibinfo{booktitle}{Advances in neural information
  processing systems}}, pp. \bibinfo{pages}{2164--2172}.

\bibitemdeclare{article}{somani2013despot}
\bibitem{somani2013despot}
\bibinfo{author}{Adhiraj \surnamestart Somani\surnameend}, \bibinfo{author}{Nan
  \surnamestart Ye\surnameend}, \bibinfo{author}{David \surnamestart
  Hsu\surnameend} \& \bibinfo{author}{Wee~Sun \surnamestart Lee\surnameend}
  (\bibinfo{year}{2013}): \emph{\bibinfo{title}{DESPOT: Online POMDP planning
  with regularization}}.
\newblock {\slshape \bibinfo{journal}{Advances in neural information processing
  systems}} \bibinfo{volume}{26}.

\bibitemdeclare{manual}{aosOverviewVideo}
\bibitem{aosOverviewVideo}
\bibinfo{author}{Dan~R. \surnamestart Suissa\surnameend}
  (\bibinfo{year}{2022}): \emph{\bibinfo{title}{A short AOS overview video}}.
\newblock \urlprefix\url{{https://www.youtube.com/watch?v=8pqZADVBLPM}}.

\bibitemdeclare{inproceedings}{sarsop}
\bibitem{sarsop}
\bibinfo{author}{David~Hsu \surnamestart Wee Sun Lee
  Hanna~Kurniawati\surnameend} (\bibinfo{year}{2008}):
  \emph{\bibinfo{title}{{SARSOP}: Efficient Point-Based {POMDP} Planning by
  Approximating Optimally Reachable Belief Spaces}}.
\newblock In: {\slshape \bibinfo{booktitle}{Proceedings of Robotics: Science
  and Systems IV}}, \bibinfo{address}{Zurich, Switzerland}, pp.
  \bibinfo{pages}{5427--5433}, \doi{10.15607/RSS.2008.IV.009}.

\bibitemdeclare{unpublished}{RDDLSimVsAOS}
\bibitem{RDDLSimVsAOS}
\bibinfo{author}{Or~\surnamestart Wertheim\surnameend} (\bibinfo{year}{2010}):
  \emph{\bibinfo{title}{An experiment comparing the generative model sampling
  rate of RDDLSim's generic code vs. the AOS's domain-specific auto-generated
  code.}}
\newblock \bibinfo{note}{Https://github.com/ssanner/rddlsim}.

\bibitemdeclare{manual}{aosArmadilloExp}
\bibitem{aosArmadilloExp}
\bibinfo{author}{Or~\surnamestart Wertheim\surnameend} (\bibinfo{year}{2021}):
  \emph{\bibinfo{title}{Armadillo experiment, detailed description}}.
\newblock
  \urlprefix\url{{https://github.com/orhaimwerthaim/AOS-experiments/tree/main/armadillo_pick}}.

\bibitemdeclare{manual}{rosplanExp}
\bibitem{rosplanExp}
\bibinfo{author}{Or~\surnamestart Wertheim\surnameend} (\bibinfo{year}{2021}):
  \emph{\bibinfo{title}{ROSPlan PDDL experiment}}.
\newblock
  \urlprefix\url{{https://github.com/orhaimwerthaim/AOS-OtherSystems-ROSPlanExperimentPDDL/}}.

\bibitemdeclare{manual}{aosArmadilloVideo}
\bibitem{aosArmadilloVideo}
\bibinfo{author}{Or~\surnamestart Wertheim\surnameend} (\bibinfo{year}{2022}):
  \emph{\bibinfo{title}{AOS Armadillo robot experiment video}}.
\newblock \urlprefix\url{{https://youtu.be/10sTQ8a_N6c}}.

\bibitemdeclare{manual}{aosExp}
\bibitem{aosExp}
\bibinfo{author}{Or~\surnamestart Wertheim\surnameend} (\bibinfo{year}{2022}):
  \emph{\bibinfo{title}{The AOS experiments documentation files}}.
\newblock \urlprefix\url{{https://github.com/orhaimwerthaim/AOS-experiments}}.

\bibitemdeclare{manual}{aosPandaExpVideo}
\bibitem{aosPandaExpVideo}
\bibinfo{author}{Or~\surnamestart Wertheim\surnameend} (\bibinfo{year}{2022}):
  \emph{\bibinfo{title}{The AOS Franka Emika Panda CoBot robot experiment
  video}}.
\newblock \urlprefix\url{{https://www.youtube.com/watch?v=-2qN4WXdvj4}}.

\bibitemdeclare{manual}{aosPandaLostExpVideo}
\bibitem{aosPandaLostExpVideo}
\bibinfo{author}{Or~\surnamestart Wertheim\surnameend} (\bibinfo{year}{2022}):
  \emph{\bibinfo{title}{The AOS Panda CoBot experiment video: the robot
  sometimes loses due to an inaccurate opponent model.}}
\newblock \urlprefix\url{{https://www.youtube.com/watch?v=R4dBrP7SLe8}}.

\bibitemdeclare{article}{younes2004ppddl1}
\bibitem{younes2004ppddl1}
\bibinfo{author}{H{\r a}kan~LS \surnamestart Younes\surnameend} \&
  \bibinfo{author}{Michael~L \surnamestart Littman\surnameend}
  (\bibinfo{year}{2004}): \emph{\bibinfo{title}{PPDDL1. 0: An extension to PDDL
  for expressing planning domains with probabilistic effects}}.
\newblock {\slshape \bibinfo{journal}{Techn. Rep. CMU-CS-04-162}}
  \bibinfo{volume}{2}, p.~\bibinfo{pages}{99}.

\end{thebibliography}

\vfill

\end{document}